\documentclass{article}

\usepackage{times}
\usepackage{paralist}
\usepackage{latexsym}
\usepackage{multirow}
\usepackage{amsthm}
\usepackage{enumitem}
\usepackage{todonotes}
\usepackage{xspace}
\usepackage[export]{adjustbox}
\usepackage{amsmath,amssymb}
\usepackage{xcolor}
\DeclareMathOperator{\E}{\mathbb{E}}
\DeclareMathOperator{\ELBO}{\mathrm{ELBO}}
\DeclareMathOperator{\KL}{\mathbb{D}_{\rm KL} }
\DeclareMathOperator{\HH}{\mathcal{H}}
\DeclareMathOperator{\II}{\mathcal{I}}
\DeclareMathOperator{\LL}{\mathcal{L}}

\newcommand \NOTE[1]{{\bf \color{red} \\NOTE:#1.\\}}

\newcommand{\dgmvae}{{DGM-VAE}\xspace} 
\newcommand{\lsebm}{{SVEBM}\xspace} 
\newcommand{\ibebm}{{SVEBM-IB}\xspace} 
\newcommand{\lsebmfull}{{Symbol-Vector Coupling Energy-Based Model}\xspace} 

\usepackage{microtype}
\usepackage{graphicx}
\usepackage{booktabs} 
\usepackage{caption}
\usepackage{subcaption}

\usepackage{hyperref}

\usepackage[accepted]{icml2021}

\icmltitlerunning{Latent Space Symbol-Vector Coupling for Text Modeling}

\begin{document}

\twocolumn[
\icmltitle{Latent Space Energy-Based Model of Symbol-Vector Coupling for Text Generation and Classification}



\icmlsetsymbol{equal}{*}

\begin{icmlauthorlist}
\icmlauthor{Bo Pang}{ucla}
\icmlauthor{Ying Nian Wu}{ucla}
\end{icmlauthorlist}

\icmlaffiliation{ucla}{Department of Statistics, University of California, Los Angeles, California, USA}

\icmlcorrespondingauthor{Bo Pang}{bopang@ucla.edu}

\icmlkeywords{Energy-Based Model, Text Modeling}

\vskip 0.3in
]



\printAffiliationsAndNotice{} 

\begin{abstract}
We propose a latent space energy-based prior model for text generation and classification. The model stands on a generator network that generates the text sequence based on a continuous latent vector. The energy term of the prior model couples a continuous  latent vector and a symbolic one-hot vector, so that discrete category can be inferred from the observed example based on the continuous latent vector. Such a latent space coupling naturally enables incorporation of information bottleneck regularization to encourage the continuous latent vector to extract information from the observed example that is informative of the underlying category.  In our learning method, the symbol-vector coupling, the generator network and the inference network are learned jointly. Our model can be learned in an unsupervised setting where no category labels are provided. It can also be learned in semi-supervised setting where category labels are provided for a subset of training examples. Our experiments demonstrate that the proposed model learns well-structured and meaningful latent space, which (1) guides the generator to generate text with high quality, diversity, and interpretability, and (2) effectively classifies text.
\end{abstract}

\section{Introduction}

Generative models for text generation is of vital importance in a wide range of real world applications such as dialog system \cite{young2013pomdp} and machine translation \citep{brown1993mathematics}. Impressive progress has been achieved with the development of neural generative models \cite{serban2016building, zhao2017learning, zhao2018unsupervised, zhang-etal-2016-variational-neural, li2017deep, gupta2018deep, zhao2018adversarially} .  However, most of prior methods focus on the improvement of text generation quality such as fluency and diversity. Besides the quality, the interpretability or controllability of text generation process is also critical for real world applications. Several recent papers recruit deep latent variable models for interpretable text generation where the latent space is learned to capture interpretable structures such as topics and dialog actions which are then used to guide text generation \citep{wang2019topic, zhao2018unsupervised}. 

Deep latent variable models map a latent vector to the observed example such as a piece of text. Earlier methods \cite{kingma2013auto, rezende2014stochastic, bowman-etal-2016-generating} utilize a continuous latent space. Although it is able to generate text of high quality, it is not suitable for modeling interpretable discrete attributes such as topics and dialog actions. A recent paper \citep{zhao2018unsupervised} proposes to use a discrete latent space in order to capture dialog actions and has shown promising interpretability of dialog utterance generation. A discrete latent space nevertheless encodes limited information and thus might limit the expressiveness of the generative model. To address this issue, \citet{shi2020dispersed} proposes to use Gaussian mixture VAE (variational auto-encoder) which has a latent space with both continuous and discrete latent variables. By including a dispersion term to avoid the modes of the Gaussian mixture to collapse into a single mode, the model produces promising results on interpretable generation of dialog utterances. 

To improve the expressivity of the latent space and the generative model as a whole, \citet{pang2020learning} recently proposes to learn an energy-based model (EBM) in the latent space, where the EBM serves as a prior model for the latent vector. Both the EBM prior and the generator network are learned jointly by maximum likelihood or its approximate variants. The latent space EBM has been applied to text modeling, image modeling, and molecule generation, and significantly improves over VAEs with Gaussian prior, mixture prior and other flexible priors. \citet{aneja2020ncpvae} generalizes this model to a multi-layer latent variable model with a large-scale generator network and achieves state-of-the-art generation performance on images.

Moving EBM from data space to latent space allows the EBM to stand on an already expressive generator model, and the EBM prior can be considered a correction of the non-informative uniform or isotropic Gaussian prior of the generative model. Due to the low dimensionality of the latent space, the EBM can be parametrized by a very small network, and yet it can capture regularities and rules in the data effectively (and implicitly).

In this work, we attempt to leverage the high expressivity of EBM prior for text modeling and learn a well-structured latent space for both interpretable generation and text classification. Thus, we formulate a new prior distribution which couples continuous latent variables (i.e., vector) for generation and discrete latent variables (i.e., symbol) for structure induction. We call our model \lsebmfull (\lsebm). 

Two key differences of our work from \citet{pang2020learning} enable incorporation of information bottleneck \citep{tishby2000information}, which encourages the continuous latent vector to extract information from the observed example that is informative of the underlying structure. First, unlike \citet{pang2020learning} where the posterior inference is done with short-run MCMC sampling, we learn an amortized inference network which can be conveniently optimized. Second, due to the coupling formulation of the continuous latent vector and the symbolic one-hot vector, given the inferred continuous vector, the symbol or category can be inferred from it via a standard softmax classifier (see Section \ref{sec:svc} for more details). 
The model can be learned in unsupervised setting where no category labels are provided. The symbol-vector coupling, the generator network, and the inference network are learned jointly by maximizing the variational lower bound of the log-likelihood. 
The model can also be learned in semi-supervised setting where the category labels are provided for a subset of training examples. The coupled symbol-vector allows the learned model to generate text from the latent vector controlled by the symbol. Moreover, text classification can be accomplished by inferring the symbol based on the continuous vector that is inferred from the observed text.

\textbf{Contributions. \quad} (1) We propose a symbol-vector coupling EBM in the latent space, which is capable of both unsupervised and semi-supervised learning. (2) We develop a regularization of the model based on the information bottleneck principle. (3) Our experiments demonstrate that the proposed model learns well-structured and meaningful latent space, allowing for interpretable text generation and effective text classification.

\begin{figure}[tb]
	\centering	
	\includegraphics[width=.52\linewidth]{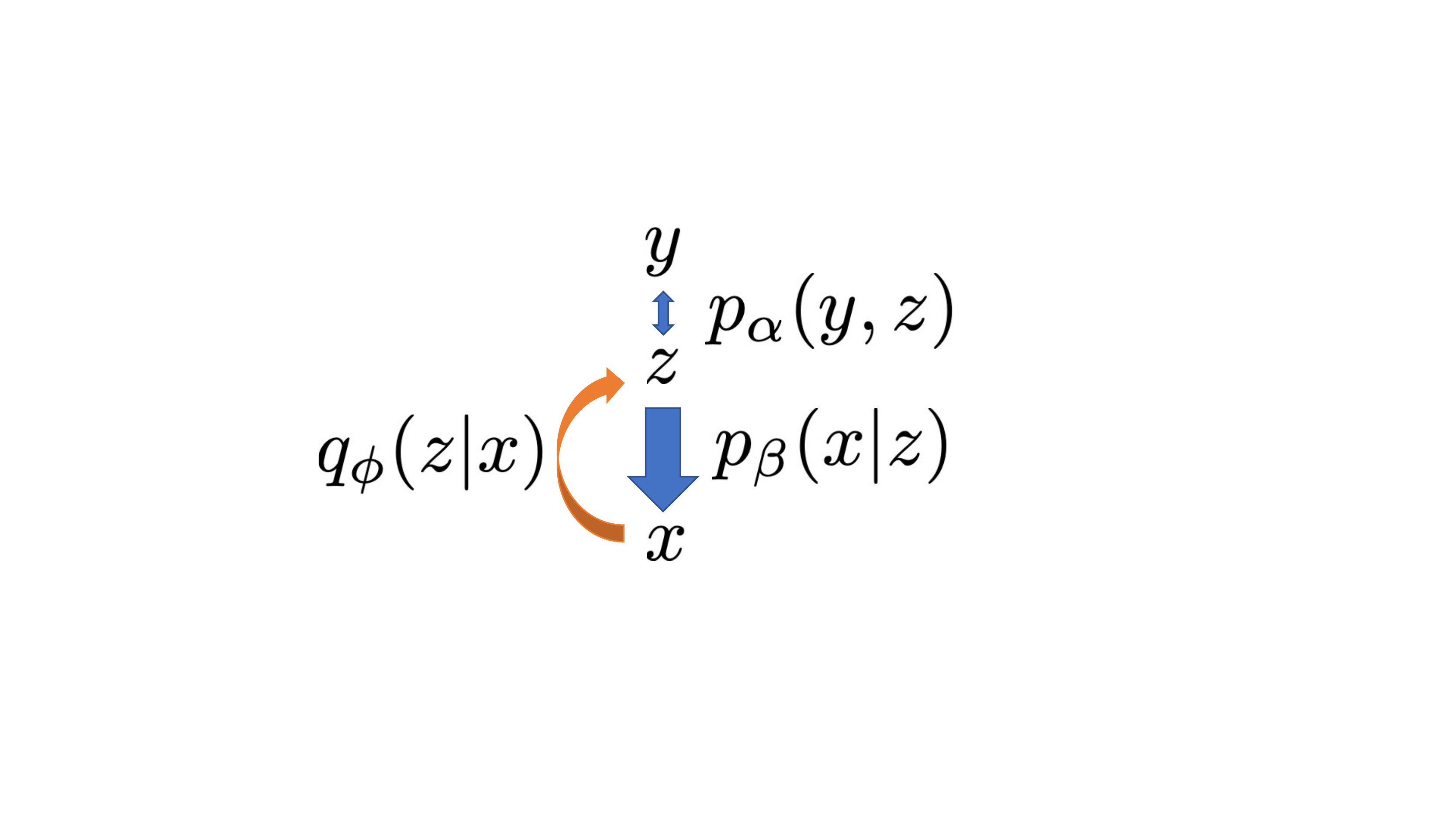}\\
	\caption{Graphical illustration of \lsebmfull (\lsebm). $y$ is a symbolic one-hot vector, and $z$ is a dense continuous vector. $x$ is the observed example. $y$ and $z$ are coupled together through an EBM, $p_\alpha(y, z)$, in the latent space. Given $z$, $y$ and $x$ are independent, i.e., $z$ is sufficient for $y$, hence giving the generator model $p_\beta(x|z)$. The intractable posterior, $p_\theta(z|x)$ with $\theta = (\alpha, \beta)$, is approximated by a variational inference model, $q_\phi(z|x)$.}
	\label{fig:ibebm}
\end{figure}

\section{Model and learning}
\subsection{Model: symbol-vector coupling} 
\label{sec:svc}
Let $x$ be the observed text sequence. Let $z \in \mathbb{R}^d$ be the continuous latent vector. Let $y$ be the symbolic one-hot vector indicating one of $K$ categories. Our generative model is defined by 
\begin{align}
    p_\theta(y, z, x) = p_\alpha(y, z) p_\beta(x | z), \label{eq:whole}
\end{align}
where $p_{\alpha}(y, z)$ is the prior model with parameters $\alpha$, $p_\beta(x|z)$ is the top-down generation model with parameters $\beta$, and $\theta = (\alpha, \beta)$. Given $z$, $y$ and $x$ are independent, i.e., $z$ is sufficient for $y$. 

The prior model $p_{\alpha}(y, z)$ is formulated as an energy-based model,
\begin{align}
p_{\alpha}(y, z) = \frac{1}{Z_\alpha} \exp(\langle y, f_\alpha(z)\rangle) p_0(z), \label{eq:prior}
\end{align}   
where $p_0(z)$ is a reference distribution, assumed to be isotropic Gaussian (or uniform) non-informative prior of the conventional generator model. $f_{\alpha}(z) \in \mathbb{R}^K$ is parameterized by a small multi-layer perceptron. $Z_\alpha$ is the normalizing constant or partition function. 

The energy term $ \langle y, f_\alpha(z)\rangle$ in Equation (\ref{eq:prior}) forms an associative memory that couples the symbol $y$ and the dense vector $z$. Given $z$, 
\begin{align} 
p_\alpha(y|z) \propto \exp(\langle y, f_\alpha(z)\rangle),  \label{eq:softmax}
\end{align}
 i.e., a softmax classifier, where $f_\alpha(z)$ provides the $K$ logit scores for the $K$ categories. Marginally, 
 \begin{align}
 p_\alpha(z) = \frac{1}{Z_\alpha}  \exp(F_\alpha(z))p_0(z),
 \end{align}
  where the marginal energy term
 \begin{align} 
 F_\alpha(z) = \log \sum_y \exp(\langle y, f_\alpha(z)\rangle), \label{eq:lse}
 \end{align} 
  i.e., the so-called log-sum-exponential form.  The summation can be easily computed because we only need to sum over $K$ different values of the one-hot $y$.

The above prior model $p_\alpha(y, z)$ stands on a generation model $p_\beta(x|z)$. For text modeling, let $x = (x^{(t)}, t=1,...,T)$ where $x^{(t)}$ is the $t$-th token. Following previous text VAE model \cite{bowman-etal-2016-generating}, we define $p_{\beta}(x|z)$ as a conditional autoregressive model,
\begin{align} 
    p_\beta(x|z) = \prod_{t=1}^T p_\beta(x^{(t)}|x^{(1)}, ..., x^{(t-1)}, z) 
\end{align} 
which is parameterized by a recurrent network with parameters $\beta$. See Figure~\ref{fig:ibebm} for a graphical illustration of our model.

\subsection{Prior and posterior sampling: symbol-aware continuous vector computation} 

Sampling from the prior $p_\alpha(z)$ and the posterior $p_\theta(z|x)$ can be accomplished by Langevin dynamics. For prior sampling from $p_\alpha(z)$, Langevin dynamics iterates 
\begin{align}
z_{t+1} = z_t + s \nabla_z \log p_\alpha(z_t) + \sqrt{2s} e_t, 
\end{align}
where $e_t \sim \mathcal{N}(0, I_d)$, $s$ is the step size, and the gradient is computed by 
\begin{align} 
\nabla_z \log p_\alpha(z) &= \E_{p_\alpha(y|z)}[\nabla_z \log p_\alpha(y, z)] \nonumber \\
&=  \E_{p_\alpha(y|z)}[\langle y, \nabla_z f_\alpha(z)\rangle ] , 
\end{align} 
where the gradient computation involves averaging $\nabla_z f_\alpha(z)$ over the softmax classification probabilities $p_\alpha(y|z)$ in Equation (\ref{eq:softmax}). Thus the sampling of the continuous dense vector $z$ is aware of the symbolic $y$. 

Posterior sampling from $p_\theta(z|x)$ follows a similar scheme, where 
\begin{align} 
\nabla_z \log p_\theta(z|x) &= \E_{p_\alpha(y|z)}[\langle y, \nabla_z f_\alpha(z)\rangle ] \nonumber \\
&+ \nabla_z \log p_\beta(x|z).
\end{align} 
When the dynamics is reasoning about $x$ by sampling the dense continuous vector $z$ from $p_\theta(z|x)$, it is aware of the symbolic $y$ via the softmax $p_\alpha(y|z)$. 

Thus $(y, z)$ forms a coupling between symbol and dense vector, which gives the name of our model, Symbol-Vector Coupling Energy-Based Model (SVEBM). 

\citet{pang2020learning} proposes to use prior and posterior sampling for maximum likelihood learning. Due to the low-dimensionality of the latent space, MCMC sampling is affordable and mixes well.

\subsection{Amortizing posterior sampling and variational learning}

Comparing prior and posterior sampling, prior sampling is particularly affordable, because $f_\alpha(z)$ is a small network. In comparison, $\nabla_z \log p_\beta(x|z)$ in the posterior sampling requires back-propagation through the generator network, which can be more expensive. Therefore we shall amortize the posterior sampling from $p_\theta(z|x)$ by an inference network, and we continue to use MCMC for prior sampling. 

Specifically, following VAE  \cite{kingma2013auto}, we recruit an inference network $q_\phi(z|x)$ to approximate the true posterior $p_\theta(z|x)$, in order to amortize posterior sampling. Following VAE, we learn the inference model $q_\phi(z|x)$ and the top-down model $p_\theta(y, z, x)$ in Equation (\ref{eq:whole}) jointly. 

For unlabeled $x$, the log-likelihood $\log p_\theta(x)$ is lower bounded by the evidence lower bound (ELBO),
\begin{align}
    &\ELBO(x|\theta,\phi) = \log p_\theta(x) - \KL(q_\phi(z|x)\|p_\theta(z|x)) \nonumber \\
    &=  \E_{q_\phi(z|x)} \left[\log p_\beta(x|z)\right] - \KL(q_\phi(z|x) \| p_\alpha(z)), 
\end{align}
where $\KL$ denotes the Kullback-Leibler divergence. 

For the prior model, the learning gradient  is 
\begin{align}
    \nabla_\alpha \ELBO = \E_{q_\phi(z|x)} [\nabla_\alpha F_{\alpha}(z)] 
    - \E_{p_\alpha(z)} [\nabla_\alpha F_{\alpha}(z)],
\end{align}
where $F_\alpha(z)$ is defined by (\ref{eq:lse}), $\E_{q_\phi(z|x)}$ is approximated by samples from the inference network, and $\E_{p_\alpha(z)}$ is approximated by persistent MCMC samples from the prior.

Let $\psi = \{\beta, \phi\}$ collect the parameters of the inference (encoder) and generator (decoder) models. The learning gradients for the two models are
\begin{align}
    &\nabla_\psi \ELBO = \nabla_\psi \E_{q_\phi(z|x)} [\log p_\beta(x|z)] \nonumber\\
    &- \nabla_\psi \KL(q_\phi(z|x) \| p_0(z)) + \nabla_\psi \E_{q_{\phi(z|x)}} [F_{\alpha}(z)], \label{eq:extra}
\end{align}
where $p_0(z)$ is the reference distribution in Equation (\ref{eq:prior}), and $\KL(q_\phi(z|x) \| p_0(z))$ is tractable. The expectations in the other two terms are approximated by samples from the inference network $q_\phi(z|x)$ with reparametrization trick  \cite{kingma2013auto}. Compared to the original VAE, we only need to include the extra $F_\alpha(z)$ term in Equation (\ref{eq:extra}), while $\log Z_\alpha$ is a constant that can be discarded. This expands the scope of VAE where the top-down model is a latent EBM. 

As mentioned above, we shall not amortize the prior sampling from $p_\alpha(z)$ due to its simplicity. Sampling $p_\alpha(z)$ is only needed in the training stage, but is not required in the testing stage.

\subsection{Two joint distributions}

Let $q_{\rm data}(x)$ be the data distribution that generates $x$. For variational learning, we maximize the averaged ELBO: $\E_{q_{\rm data}(x)}[\ELBO(x|\theta,\phi)]$, where $\E_{q_{\rm data}(x)}$ can be approximated by averaging over the training examples. Maximizing $\E_{q_{\rm data}(x)}[\ELBO(x|\theta,\phi)]$ over $(\theta, \phi)$ is equivalent to minimizing the following objective function over $(\theta, \phi)$
\begin{align}
& \KL(q_{\rm data}(x) \| p_\theta(x)) 
    +  \E_{q_{\rm data}(x)} [\KL(q_\phi(z|x) \| p_\theta(z|x))]  \nonumber \\
  &= \KL(q_{\rm data}(x) q_\phi(z|x)\| p_\alpha(z) p_\beta(x | z) ).  \label{eq:joint}
\end{align} 
The right hand side is the KL-divergence between two joint distributions: $Q_\phi(x, z) = q_{\rm data}(x) q_\phi(z|x)$, and $P_\theta(x, z) = p_\alpha(z) p_\beta(x|z)$. 
The reason we use notation $q$ for the data distribution $q_{\rm data}(x)$ is for notation consistency. Thus VAE can be considered as joint minimization of $\KL(Q_\phi\|P_\theta)$ over $(\theta, \phi)$. Treating $(x, z)$ as the complete data, $Q_\phi$ can be considered the complete data distribution, while $P_\theta$ is the model distribution of the complete data. 

For the distribution $Q_\phi(x, z)$, we can define the following quantities. 
\begin{align}
q_\phi(z) = \E_{q_{\rm data}(x)} [q_\phi(z|x)] = \int Q_\phi(x, z) dx
\end{align}
is the aggregated posterior distribution and the marginal distribution of $z$ under $Q_\phi$. $\HH(z)=-\E_{q_\phi(z)}[ \log q_\phi(z)]$ is the entropy of the aggregated posterior $q_\phi(z)$. 

$\HH(z|x) = - \E_{Q_\phi(x, z)}[ \log q_\phi(z|x)]$ is the conditional entropy of $z$ given $x$ under the variational inference distribution $q_{\phi}(z|x)$. 
\begin{align}
&\II(x, z) = \HH(z) - \HH(z|x) \nonumber \\
& = -\E_{q_\phi(z)} [\log q_\phi(z)] + \E_{Q_\phi(x, z)}[ \log q_\phi(z|x)]
\end{align}
is the mutual information between $x$ and $z$ under $Q_\phi$. 

It can be shown that the VAE objective in Equation (\ref{eq:joint}) can be written as 
\begin{align}  
\KL&(Q_\phi(x, z) \| P_\theta(x, z)) \nonumber \\
    &=-\HH(x) - \E_{Q_\phi(x, z)} [\log p_\beta(x|z)] \nonumber \\
    & \quad +\II(x, z) + \KL(q_\phi(z) \| p_\alpha(z)), \label{eq:joint1}
\end{align}
where 
 $\HH(x) = - \E_{q_{\rm data}(x)} [\log q_{\rm data}(x)]$ is the entropy of the data distribution and is fixed.

\subsection{Information bottleneck} 

Due to the coupling of $y$ and $z$ (see Equations (\ref{eq:prior}) and (\ref{eq:softmax})), a learning objective with information bottleneck can be naturally developed as a simple modification of the VAE objective in Equations (\ref{eq:joint}) and (\ref{eq:joint1}): 
\begin{align}
    \LL(\theta, \phi) &= \KL(Q_\phi(x, z)\| P_\theta(x, z)) - \lambda \II(z, y) \label{eq:ib}\\
    &=-\HH(x) - \underbrace{\E_{Q_\phi(x, z)}[ \log p_\beta(x|z)]}_\text{reconstruction} \label{eq:recon}\\
    &\quad + \underbrace{\KL(q_\phi(z) \| p_\alpha(z))}_\text{EBM learning}\\
    &\quad + \underbrace{\II(x, z) - \lambda \II(z, y)}_\text{information bottleneck},
\end{align}
where $\lambda \ge 0$ controls the trade-off between the compressivity of $z$ about $x$ and its expressivity to $y$. The mutual information between $z$ and $y$, $\II(z, y)$, is defined as:
\begin{align}
\II(z, y) &= \HH(y) - \HH(y|z) \nonumber\\
& = -\sum_y q(y) \log q(y) \nonumber\\
&\quad + \E_{q_\phi(z)} \sum_y p_\alpha(y|z) \log p_\alpha(y|z),  \label{eq:mi}
\end{align}
where $q(y) = \E_{q_\phi(z)} [p_\alpha(y|z)]$. $\II(z, y)$, $\HH(y)$, and $\HH(y|z)$ are defined based on $Q(x, y, z) = q_{\rm data}(x) q_\phi(z|x) p_\alpha(y|z)$, where $p_\alpha(y|z)$ is softmax probability over $K$ categories in Equation (\ref{eq:softmax}). 

In computing $\II(z, y)$, we need to take expectation over $z$ under $q_\phi(z) = \E_{q_{\rm data}(x)} [q_\phi(z|x)]$, which is approximated with a mini-batch of $x$ from $q_{\rm data}(x)$ and multiple samples of $z$ from $q_\phi(z|x)$ given each $x$. 

The Lagrangian form of the classical information bottleneck objective \citep{tishby2000information} is,
\begin{align}
\min_{p_\theta(z|x)} [\II(x, z|\theta) -\lambda \II(z, y|\theta)] \label{eq:cib}.
\end{align} 
Thus minimizing $\LL(\theta, \phi)$ (Equation (\ref{eq:ib})) includes minimizing a variational version (variational information bottleneck or VIB; \citealt{alemi2016deep}) of Equation (\ref{eq:cib}). We do not exactly minimize VIB due to the reconstruction term in Equation (\ref{eq:recon}) that drives unsupervised learning, in contrast to supervised learning of VIB in \citet{alemi2016deep}. 

We call the \lsebm learned with the objective incorporating information bottleneck (Equation (\ref{eq:ib})) as \ibebm.

\subsection{Labeled data}
\label{sec:ssl} 

For a labeled example $(x, y)$, the log-likelihood can be decomposed into $\log p_\theta(x, y) = \log p_\theta(x) + \log p_\theta(y |x)$. The gradient of $\log p_\theta(x)$ and its ELBO can be computed in the same way as the unlabeled data described above. 
\begin{align} 
    p_\theta(y|x) = \E_{p_\theta(z|x)} [p_\alpha(y|z)] \approx \E_{q_\phi(z|x)} [p_\alpha(y|z)] \label{eq:yx}, 
\end{align} 
where $p_\alpha(y|z)$ is the softmax classifier defined  by Equation (\ref{eq:softmax}), and $q_\phi(z|x)$ is the learned inference network. In practice, $\E_{q_\phi(z|x)} [p_\alpha(y|z)]$ is further approximated by $p_\alpha(y | z = \mu_\phi(x))$ where $\mu_\phi(x)$ is the posterior mean of $q_\phi(z|x)$. We found using $\mu_\phi(x)$ gave better empirical performance than using multiple posterior samples. 
 
 For semi-supervised learning, we can combine the learning gradients from both unlabeled and labeled data. 
 
 \subsection{Algorithm}
The learning and sampling algorithm for \lsebm is described in Algorithm \ref{alg:lsebm}. Adding the respective gradients of $\II(z, y)$ (Equation (\ref{eq:mi})) to Step 4 and Step 5 allows for learning \ibebm.

\begin{algorithm}[tb]
  \caption{Unsupervised and Semi-supervised Learning of \lsebmfull.}
  \label{alg:lsebm}
\begin{algorithmic}
  \STATE {\bfseries Input:} Learning iterations~$T$, learning rates~$(\eta_0,\eta_1,\eta_2)$, initial parameters~$(\alpha_0, \beta_0, \phi_0)$, observed unlabelled examples~$\{x_i\}_{i=1}^M$, observed labelled examples~$\{(x_i, y_i)\}_{i=M+1}^{M+N}$ (optional, needed only in semi-supervised learning), unlabelled and labelled batch sizes $(m,n)$, initializations of persistent chains~$\{z_i^{-}\sim p_0(z)\}_{i=1}^{L}$, and number of Langevin dynamics steps $T_{LD}$. 
    \STATE {\bfseries Output:} $(\alpha_{T}, \beta_{T}, \phi_{T})$.
  \FOR{$t=0$ {\bfseries to} $T-1$}
  \STATE {\bfseries 1. mini-batch:} Sample unlabelled $\{ x_i \}_{i=1}^m$ and labelled observed examples $\{ x_i, y_i \}_{i=m+1}^{m+n}$.
  \STATE {\bfseries 2. prior sampling:} For each unlabelled $x_i$, randomly pick and update a persistent chain $z_i^{-}$ by Langevin dynamics with target distribution $p_\alpha(z)$ for $T_{LD}$ steps.   
  \STATE {\bfseries 3. posterior sampling:} For each $x_i$, sample $z_i^{+} \sim q_\phi(z|x_i)$ using the inference network and reparameterization trick.
  \STATE {\bfseries 4. unsupervised learning of prior model:} $\alpha_{t+1} = \alpha_t + \eta_0 \frac{1}{m}\sum_{i=1}^{m} [\nabla_\alpha F_{\alpha_t}(z_i^{+}) - \nabla_\alpha F_{\alpha_t}(z_i^{-})]$. 
  \STATE {\bfseries 5. unsupervised learning of inference and generator models:} \\$\psi_{t+1} = \psi_t + \eta_1 \frac{1}{m}\sum_{i=1}^{m}[\nabla_\psi \log p_{\beta_t}(x_i|z_i^{+}) - \nabla_\psi \KL(q_{\phi_t}(z|x_i) \| p_0(z)) + \nabla_\psi F_{\alpha_t}(z_i^{+})]$, with backpropagation through $z_i^{+}$ via reparametrization trick.\\ 

  \IF{labeled examples $(x, y)$ are available}
  \STATE {\bfseries 6. supervised learning of prior and inference models:}  Let $\gamma = (\alpha, \phi)$. 
  $\gamma_{t+1} = \gamma_t + \eta_2 \frac{1}{n}\sum_{i=m+1}^{m+n}$ \\$\nabla_\gamma \log p_{\alpha_t}(y_i| z_i= \mu_{\phi_t}(x_i))$.
  \ENDIF
  \ENDFOR
\end{algorithmic}
\end{algorithm}

\section{Experiments}
We present a set of experiments to assess (1) the quality of text generation, (2) the interpretability of text generation, and (3) semi-supervised classification of our proposed models, \lsebm and \ibebm, on standard benchmarks. The proposed \lsebm is highly expressive for text modeling and demonstrate superior text generation quality and is able to discover meaningful latent labels when some supervision signal is available, as evidenced by good semi-supervised classification performance. \ibebm not only enjoys the expressivity of \lsebm but also is able to discover meaningful labels in an unsupervised manner since the information bottleneck objective encourages the continuous latent variable, $z$, to keep sufficient information of the observed $x$ for the emergence of the label, $y$.  Its advantage is still evident when supervised signal is provided.

\begin{figure}
    \centering
	\begin{subfigure}[b]{0.35\textwidth}
    \centering
		\includegraphics[width=\textwidth]{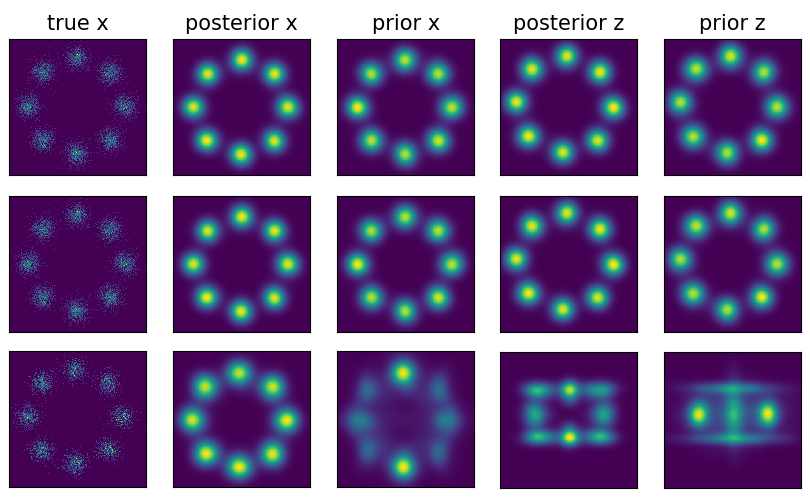}
	\end{subfigure}
	\hfill
	\hfill
	\hfill
	\begin{subfigure}[b]{0.35\textwidth}
    \centering
		\includegraphics[width=\textwidth]{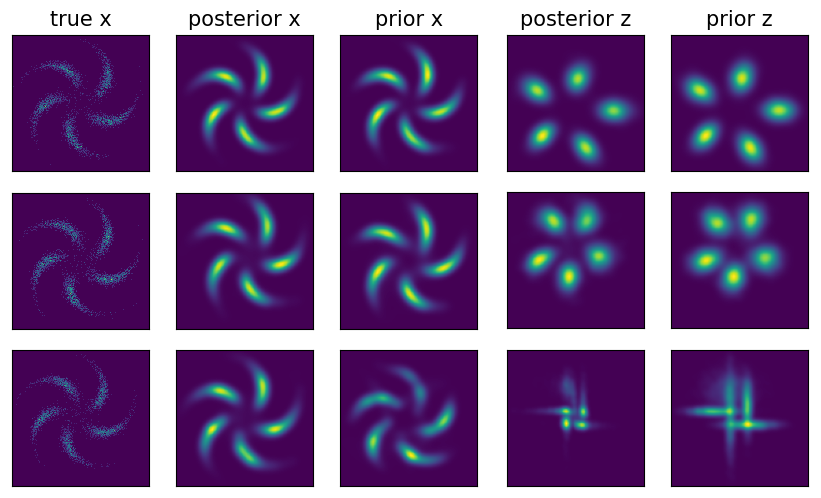}
	\end{subfigure}
	\caption{Evaluation on 2D synthetic data: a mixture of eight Gaussians (upper panel) and a pinwheel-shaped distribution (lower panel). In each panel, the first, second, and third row display densities learned by \ibebm, \lsebm, and \dgmvae, respectively.} 
	\label{fig:8gaussians}
\end{figure}

\subsection{Experiment settings} 
Generation quality is evaluated on the Penn Treebanks (\citealt{marcus1993building}, PTB) as pre-processed by \citet{mikolov2010recurrent}. Interpretability is first assessed on two dialog datasets, the Daily Dialog dataset \citep{li2017dailydialog} and the Stanford Multi-Domain Dialog (SMD) dataset \citep{eric2017key-value}. DD is a chat-oriented dataset and consists of $13,118$ daily conversations for English learner in a daily life. It provides human-annotated dialog actions and emotions for the utterances. SMD has $3,031$ human-Woz, task-oriented dialogues collected from three different domains (navigation, weather, and scheduling). We also evaluate generation interpretability of our models on sentiment control with Yelp reviews, as preprocessed by \citet{li-etal-2018-delete}. It is on a larger scale than the aforementioned datasets, and contains $180,000$ negative reviews and $270,000$ positive reviews. 

Our model is compared with the following baselines: (1) RNNLM \citep{mikolov2010recurrent}, language model implemented with GRU \citep{cho2014learning}; (2) AE \citep{vincent2010stacked}, deterministic autoencoder which has no regularization to the latent space; (3) DAE, autoencoder with a discrete latent space; (4) VAE \citep{kingma2013auto}, the vanilla VAE with a continuous latent space and a Gaussian noise prior; (5) DVAE, VAE with a discrete latent space; (6) DI-VAE \citep{zhao2018unsupervised}, a DVAE variant with a mutual information term between $x$ and $z$; (7) semi-VAE \citep{kingma2014semi}, semi-supervised VAE model with independent discrete and continuous latent variables; (8) GM-VAE, VAE with discrete and continuous latent variables following a Gaussian mixture; (9) DGM-VAE \citep{shi2020dispersed}, GM-VAE with a dispersion term which regularizes the modes of Gaussian mixture to avoid them collapsing into a single mode; (10) semi-VAE $+ \II(x,y)$, GM-VAE $+ \II(x,y)$, DGM-VAE $+ \II(x,y)$, are the same models as (7), (8), and (9) respectively, but with an mutual information term between $x$ and $y$ which can be computed since they all learn two separate inference networks for $y$ and $z$. To train these models involving discrete latent variables, one needs to deal with the non-differentiability of them in order to learn the inference network for $y$. In our models, we do not need a separate inference network for $y$, which can conveniently be inferred from $z$ given the inferred $z$ (see Equation \ref{eq:softmax}), and have no need to sample from the discrete variable in training. 

The encoder and decoder in all models are implemented with a single-layer GRU with hidden size 512. The dimensions for the continuous vector are $40$, $32$, $32$, and $40$ for PTB, DD, SMD and Yelp, respectively. The dimensions for the discrete variable are $20$ for PTB, $125$ for DD, $125$ for SMD, and $2$ for Yelp. $\lambda$ in information bottleneck (see Equation \ref{eq:ib}) that controls the trade-off between compressivity of $z$ about $x$ and its expressivity to $y$ is not heavily tuned and set to $50$ for all experiments. Our implementation is available at \url{https://github.com/bpucla/ibebm.git}.

\subsection{2D synthetic data}
We first evaluate our models on 2-dimensional synthetic datasets for direct visual inspection. They are compared to the best performing baseline in prior works, \dgmvae $+ \II(x,y)$ \citep{shi2020dispersed}. The results are displayed in Figure~\ref{fig:8gaussians}. In each row, \textit{true x} indicates the true data distribution $q_{\rm data}(x)$; \textit{posterior x} indicates the KDE (kernel density estimation) distribution of $x$ based on $z$ samples from its posterior $q_\phi(z|x)$; \textit{prior x} indicates the KDE of $p_\theta(x) = \int p_\beta(x|z) p_\alpha(z) dz$, based on $z$ samples from the learned EBM prior, $p_\alpha(z)$; \textit{posterior z} indicates the KDE of the aggregate posterior, $q_\phi(z) = \int q_{\rm data}(x)q_\phi(z|x) dx$; \textit{prior z} indicates the KDE of the learned EBM prior, $p_\alpha(z)$.

It is clear that our proposed models, \lsebm and \ibebm model the data well in terms of both \textit{posterior x} and \textit{prior x}. In contrast, although \dgmvae reconstructs the data well but the learned generator $p_\theta(x)$ tend to miss some modes. The learned prior $p_\theta(z)$ in \lsebm and \ibebm shows the same number of modes as the data distribution and manifests a clear structure. Thus, the well-structured latent space is able to guide the generation of $x$. By comparison, although \dgmvae shows some structure in the latent space, the structure is less clear than that of our model. It is also worth noting that \lsebm performs similarly as \ibebm, and thus the symbol-vector coupling \textit{per se}, without the information bottleneck, is able to capture the latent space structure of relatively simple synthetic data. 

\subsection{Language generation}
We evaluate the quality of text generation on PTB and report four metrics to assess the generation performance: reverse perplexity (rPPL; \citealt{zhao2018adversarially}), BELU \cite{papineni2002bleu}, word-level KL divergence (wKL), and negative log-likelihood (NLL). Reverse perplexity is the perplexity of ground-truth test set computed under a language model trained with generated data. Lower rPPL indicates that the generated sentences have higher diversity and fluency. We recruit ASGD Weight-Dropped LSTM \cite{merity2018regularizing}, a well-performed and popular language model, to compute rPPL. The synthesized sentences are sampled with $z$ samples from the learned latent space EBM prior, $p_\alpha(z)$. The BLEU score is computed between the input and reconstructed sentences and measures the reconstruction quality. Word-level KL divergence between the word frequencies of training data and synthesized data reflects the generation quality. Negative log-likelihood~\footnote{It is computed with importance sampling \cite{burda2015importance} with 500 importance samples.} measures the general model fit to the data. These metrics are evaluated on the test set of PTB, except wKL, which is evaluated on the training set. 

The results are summarised in Table~\ref{tab:generation_quality}. Compared to previous models with (1) only continuous latent variables, (2) only discrete latent variables, and (3) both discrete and continuous latent variables, the coupling of discrete and continuous latent variables in our models through an EBM is more expressive. The proposed models, \lsebm and \ibebm, demonstrate better reconstruction (higher BLEU) and higher model fit (lower NLL) than all baseline models except AE. Its sole objective is to reconstruct the input and thus it can reconstruct sentences well but cannot generate diverse sentences. 

The expressivity of our models not only allows for capturing the data distribution well but also enables them to generate sentences of high-quality. As indicated by the lowest rPPL, our models improve over these strong baselines on fluency and diversity of generated text. Moreover, the lowest wKL of our models indicate that the word distribution of the generated sentences by our models is most consistent with that of the data.

It is worth noting that \lsebm and \ibebm have close performance on language modeling and text generation. Thus the mutual information term does not lessen the model expressivity.

\begin{table}[!htbp]
\scriptsize
\centering
\begin{tabular}{l c c c c}
\toprule
{\bf Model} & {\bf rPPL$^\downarrow$} & {\bf BLEU$^\uparrow$} & {\bf wKL$^\downarrow$} & {\bf NLL$^\downarrow$} \\
\midrule
{ Test Set} & - & 100.0 & \textbf{0.14} & -\\ 
\midrule
{ RNN-LM} & -  & - & - & 101.21 \\
\midrule 
{ AE} & 730.81 & {\bf 10.88} & 0.58 & - \\ 
{ VAE} & 686.18 & 3.12 & 0.50 & 100.85 \\ 
\midrule
{ DAE } & 797.17 & 3.93 & 0.58 & - \\ 
{ \textsc{DVAE}} & 744.07 & 1.56 & 0.55  & 101.07 \\ 
{ \textsc{DI-VAE}} & 310.29 & 4.53 & 0.24 & 108.90 \\ 
\midrule
{ semi-VAE} & 494.52 & 2.71 & 0.43 & 100.67  \\ 
{ semi-VAE $+ \II(x,y)$} & 260.28 & 5.08 & 0.20 & 107.30 \\ 
{ \textsc{GM-VAE}} & 983.50 & 2.34 & 0.72 & 99.44 \\ 
{ \textsc{GM-VAE} $+ \II(x,y)$} & 287.07 & 6.26 & 0.25 & 103.16 \\ 
{ \dgmvae } & 257.68 & 8.17 & 0.19  & 104.26 \\ 
{ \dgmvae $+ \II(x,y)$} & 247.37 & 8.67 & 0.18 & 105.73 \\ 
\midrule
{ \lsebm } & 180.71 & \textbf{9.54} & 0.17 & 95.02 \\
{ \ibebm } & \textbf{177.59} & 9.47 & \textbf{0.16} & \textbf{94.68} \\
\bottomrule
\end{tabular}
\caption{ Results of language generation on PTB.}\label{tab:generation_quality} 
\end{table}

\subsection{Interpretable generation}
We next turn to evaluate our models on the interpretabiliy of text generation.

\textbf{Unconditional text generation.\quad} The dialogues are flattened for unconditional modeling. Utterances in DD are annotated with action and emotion labels. The generation interpretability is assessed through the ability to unsupervisedly capture the utterance attributes of DD. The label, $y$, of an utterance, $x$, is inferred from the posterior distribution, $p_\theta(y|x)$ (see Equation \ref{eq:yx}). In particular, we take $y = {\rm argmax}_k p_\theta(y=k|x)$ as the inferred label. As in \citet{zhao2018unsupervised} and \citet{shi2020dispersed}, we recruit homogeneity to evaluate the consistency between groud-truth action and emotion labels and those inferred from our models. Table~\ref{tab:dd} displays the results of our models and baselines. 
Without the mutual information term to encourage $z$ to retain sufficient information for label emergence, the continuous latent variables in \lsebm appears to mostly encode information for reconstructing $x$ and performs the best on sentence reconstruction. However, the encoded information in $z$ is not sufficient for the model to discover interpretable labels and demonstrates low homogeneity scores. In contrast, \ibebm is designed to encourage $z$ to encode information for an interpretable latent space and greatly improve the interpretability of text generation over \lsebm and models from prior works, as evidenced in the highest homogeneity scores on action and emotion labels.

\begin{table}[!htbp]
    \scriptsize
    \centering
    \hspace{-2 pt}
     \begin{tabular}{l c c c c}
    \toprule
    {\bf Model}  & {\bf MI$^\uparrow$} & {\bf BLEU$^\uparrow$} & {\bf Action$^\uparrow$} & {\bf Emotion$^\uparrow$} \\
    \midrule
    { \text{DI-VAE}} &   1.20 & 3.05 & 0.18 & 0.09\\
    \midrule
    { semi-VAE} &  0.03 & 4.06 & 0.02 & 0.08\\ 
    { semi-VAE $+ \II(x,y)$} &   1.21 & 3.69 & 0.21 & 0.14\\ 
    { \textsc{GM-VAE}} &   0.00 & 2.03 & 0.08 & 0.02\\
    { \textsc{GM-VAE} $+ \II(x,y)$}  &  1.41 & 2.96 & 0.19 & 0.09\\
    { \dgmvae } &   0.53 & 7.63 & 0.11 & 0.09\\ 
    { \dgmvae $+ \II(x,y)$} & 1.32 & 7.39 &  0.23  & 0.16 \\
    \midrule
    { \lsebm} & 0.01 & {\bf 11.16} &  0.03  &  0.01 \\
    { \ibebm} & {\bf 2.42} & 10.04 & {\bf 0.59}  & {\bf 0.56} \\
    \bottomrule
    \end{tabular}
     \caption{ Results of interpretable language generation on DD. Mutual information (MI), BLEU and homogeneity with actions and emotions are shown. 
     }\label{tab:dd}
\end{table}

\textbf{Conditional text generation.\quad} We then evaluate \ibebm on dialog generation with SMD. BELU and three word-embedding-based topic similarity metrics, embedding average, embedding extrema and embedding greedy \citep{mitchell2008vector, forgues2014bootstrapping, rus2012comparison}, are employed to evaluate the quality of generated responses. The evaluation results are summarized in Table~\ref{tab:smd_quality}. \ibebm outperforms all baselines on all metrics, indicating the high-quality of the generated dialog utterances. 

SMD does not have human annotated action labels. We thus assess \ibebm qualitatively. Table~\ref{tab:case-actions} shows dialog actions discovered by it and their corresponding utterances. The utterances with the same action are assigned with the same latent code ($y$) by our model. Table~\ref{tab:case-response} displays dialog responses generated with different values of $y$ given the same context. It shows that \ibebm is able to generate interpretable utterances given the context.  

\begin{table}[!htbp]
    \scriptsize
    \centering
     \begin{tabular}{lllll}
    \toprule
    {\bf Model} & {\bf BLEU$^\uparrow$} & {\bf Average$^\uparrow$} & {\bf Extrema$^\uparrow$} & {\bf Greedy$^\uparrow$}  \\
     \midrule
    DI-VAE & 7.06 & 76.17 & 43.98 & 60.92  \\
    \dgmvae $+ \II(x,y)$ & 10.16 & 78.93 & 48.14 & 64.87  \\
    \ibebm & \bf 12.01 & \bf 80.88 & \bf \bf 51.35 & \bf 67.12  \\
    \bottomrule
    \end{tabular}
     \caption{ Dialog evaluation results on SMD with four metrics: BLEU, average, extrema and greedy word embedding based similarity. }\label{tab:smd_quality}
\end{table}

\begin{table}[!htbp]
    \centering
    \scriptsize
    \begin{tabular}{ll}
    \toprule
    {\bf Action} & Inform-weather \\
    \midrule
    \multirow{3}{*}{\bf Utterance} & Next week it will rain on Saturday in Los Angeles \\
    & It will be between 20-30F in Alhambra on Friday. \\
    & It won't be overcast or cloudy at all this week in Carson \\
    \bottomrule
    \toprule
    {\bf Action} & Request-traffic/route \\
    \midrule
    \multirow{3}{*}{\bf Utterance} & Which one is the quickest, is there any traffic? \\
    & Is that route avoiding heavy traffic? \\
    & Is there an alternate route with no traffic? \\
    \bottomrule
    \end{tabular}
    \caption{Sample actions and corresponding utterances discovered by \ibebm on SMD. }\label{tab:case-actions} 
\end{table}

\begin{table}[t]
\centering
\scriptsize
\begin{tabular}{ll}
\toprule
\multirow{2}{*}{\bf Context} & \textit{Sys:} What city do you want to hear the forecast for?\\
&\textit{User:} Mountain View\\
\midrule
\multirow{4}{*}{\bf Predict} & Today in Mountain View is gonna be overcast, with low of 60F \\
&and high of 80F. \\
\\
&  What would you like to know about the weather for Mountain View? \\
\bottomrule
\toprule
\multirow{3}{*}{\bf Context} & \textit{User:} Where is the closest tea house? \\
&\textit{Sys:} Peets Coffee also serves tea. They are 2 miles away \\
&at 9981 Archuleta Ave.\\
\midrule
\multirow{3}{*}{\bf Predict} & OK, please give me an address and directions via the shortest distance.  \\
\\
& Thanks! \\
\bottomrule
\end{tabular}
\caption{ Dialog cases on SMD, which are generated by sampling dialog utterance $x$ with different values of $y$.}\label{tab:case-response}
\end{table}

\textbf{Sentence attribute control. \quad} We evaluate our model's ability to control sentence attribute. In particular, it is measured by the accuracy of generating sentences with a designated sentiment. This experiment is conducted with the Yelp reviews. Sentences are generated given the discrete latent code $y$. A pre-trained classifier is used to determine which sentiment the generated sentence has. The pre-trained classifier has an accuracy of $98.5\%$ on the testing data, and thus is able to accurately evaluate a sentence's sentiment. There are multiple ways to cluster the reviews into two categories or in other words the sentiment attribute is not identifiable. Thus the models are trained with sentiment supervision. In addition to \dgmvae $+ \II(x,y)$, we also compare our model to text conditional GAN \cite{subramanian2018towards}.  

The quantitative results are summarized in Table~\ref{tab:yelp-accuracy}. All models have similar high accuracies of generating positive reviews. The accuracies of generating negative reviews are however lower. This might be because of the unbalanced proportions of positive and negative reviews in the training data. Our model is able to generate negative reviews with a much higher accuracy than the baselines, and has the highest overall accuracy of sentiment control. Some generated samples with a given sentiment are displayed in Table~\ref{tab:yelp-samples}.      

\begin{table}[!htbp]
    \scriptsize
    \centering
    \hspace{-2 pt}
     \begin{tabular}{l c c c}
    \toprule
    {\bf Model}  & {\bf Overall$^\uparrow$} & {\bf Positive$^\uparrow$} & {\bf Negative$^\uparrow$}  \\
    \midrule
    { \dgmvae $+ \II(x,y)$} & 64.7\% & 95.3\% &  34.0\%  \\
    { CGAN} & 76.8\% & 94.9\% &  58.6\%  \\
    { \ibebm} & {\bf 90.1\%} & 95.1\% &  {\bf 85.2\%}  \\
    \bottomrule
    \end{tabular}
     \caption{Accuracy of sentence attribute control on Yelp. 
     }\label{tab:yelp-accuracy}
\end{table}

\begin{table}[t]
\centering
\scriptsize
\begin{tabular}{ll}
\toprule
\multirow{5}{*}{\bf Positive} & The staff is very friendly and the food is great.\\
& The best breakfast burritos in the valley.\\
& So I just had a great experience at this hotel.\\
& It's a great place to get the food and service.\\
& I would definitely recommend this place for your customers.\\
\midrule
\multirow{5}{*}{\bf Negative} & I have never had such a bad experience.\\
& The service was very poor. \\
& I wouldn't be returning to this place.\\
& Slowest service I've ever experienced. \\
& The food isn't worth the price. \\
\bottomrule
\end{tabular}
\caption{Generated positive and negative reviews with \ibebm trained on Yelp.}\label{tab:yelp-samples}
\end{table}

\subsection{Semi-supervised classification}
We next evaluate our models with supervised signal partially given to see if they can effectively use provided labels. Due to the flexible formulation of our model, they can be naturally extended to semi-supervised settings (Section \ref{sec:ssl}). 

In this experiment, we switch from neural sequence models used in previous experiments to neural document models \cite{miao2016neural, card2018neural} to validate the wide applicability of our proposed models. Neural document models use bag-of-words representations. Each document is a vector of vocabulary size and each element represents a word's occurring frequency in the document, modeled by a multinominal distribution. Due to the non-autoregressive nature of neural document model, it involves lower time complexity and is more suitable for low resources settings than neural sequence model.

We compare our models to VAMPIRE \cite{gururangan2019variational}, a recent VAE-based semi-supervised learning model for text, and its more recent variants (Hard EM and CatVAE in Table~\ref{tab:ssl}) \cite{jin-etal-2020-discrete} that improve over VAMPIRE. Other baselines are (1) supervised learning with randomly initialized embedding; (2) supervised learning with Glove embedding pretrained on $840$ billion words (Glove-OD); (3) supervised learning with Glove embedding trained on in-domain unlabeled data (Glove-ID); (4) self-training where a model is trained with labeled data and the predicted labels with high confidence is added to the labeled training set. The models are evaluated on AGNews \cite{zhang2015character} with varied number of labeled data. It is a popular benchmark for text classification and contains $127,600$ documents from $4$ classes. 

The results are summarized in Table~\ref{tab:ssl}. \lsebm has reasonable performance in the semi-supervised setting where partial supervision signal is available. \lsebm performs better or on par with Glove-OD, which has access to a large amount of out-of-domain data, and VAMPIRE, the model specifically designed for text semi-supervised learning. It suggests that \lsebm is effective in using labeled data. These results support the validity of the proposed symbol-vector coupling formation for learning a well-structured latent space. \ibebm outperforms all baselines especially when the number of labels is limited ($200$ or $500$ labels), clearly indicating the effectiveness of the information bottleneck for inducing structured latent space.

\begin{table}[!htbp]
    \scriptsize
    \centering
     \begin{tabular}{l c c c c}
    \toprule
    {\bf Model}  & {200} & {500} & {2500} & {10000} \\
    \midrule
    { \text{Supervised}} &   68.8 & 77.3 & 84.4 & 87.5\\
    { Self-training} &  77.3 & 81.3 & 84.8 & 87.7\\ 
    { Glove-ID} &   70.4 & 78.0 & 84.1 & 87.1\\ 
    { Glove-OD} &   68.8 & 78.8 & 85.3 & 88.0\\
    { VAMPIRE}  &  82.9 & 84.5 & 85.8 & 87.7\\
    { Hard EM}  &  83.9 & 84.6 & 85.1 & 86.9\\
    { CatVAE}  &  84.6 & 85.7 & 86.3 & 87.5\\
    \midrule
    { \lsebm} & 84.5 & 84.7 &  86.0  &  88.1 \\
    { \ibebm} & {\bf 86.4} & {\bf 87.4} & {\bf 87.9}  & {\bf 88.6} \\
    \bottomrule
    \end{tabular}
     \caption{ Semi-supervised classification accuracy on AGNews with varied number of labeled data. 
     }\label{tab:ssl}
\end{table}

\section{Related work and discussions}
\textbf{Text generation.\quad} VAE is a prominent generative model \cite{kingma2013auto, rezende2014stochastic}. It is first applied to text modeling by \citet{bowman-etal-2016-generating}. Following works apply VAE to a wide variety of challenging text generation problems such as dialog generation \cite{serban2016building, serban2017hierarchical, zhao2017learning, zhao2018unsupervised}, machine translation \cite{zhang-etal-2016-variational-neural}, text summarization \cite{li2017deep}, and paraphrase generation \cite{gupta2018deep}. Also, a large number of following works have endeavored to improve language modeling and text generation with VAE by addressing issues like posterior collapse \cite{zhao2018adversarially, li-etal-2019-surprisingly, fu-etal-2019-cyclical, he2018lagging}. 

Recently, \citet{zhao2018unsupervised} and \citet{shi2020dispersed} explore the interpretability of text generation with VAEs. While the model in \citet{zhao2018unsupervised} has a discrete latent space, in \citet{shi2020dispersed} the model contains both discrete ($y$) and continuous ($z$) variables which follow Gaussian mixture. Similarly, we use both discrete and continuous variables. But they are coupled together through an EBM which is more expressive than Gaussian mixture as a prior model, as illustrated in our experiments where both \lsebm and \ibebm outperform the models from \citet{shi2020dispersed} on language modeling and text generation. Moreover, our coupling formulation makes the mutual information between $z$ and $y$ can be easily computed without the need to train and tune an additional auxiliary inference network for $y$ or deal with the non-diffierentibility with regard to it, while \citet{shi2020dispersed} recruits an auxiliary network to infer $y$ conditional on $x$ to compute their mutual information \footnote{Unlike our model which maximizes the mutual information between $z$ and $y$ following the information bottleneck principle \cite{tishby2000information}, they maximizes the mutual information between the observed data $x$ and the label $y$.}. \citet{kingma2014semi} also proposes a VAE with both discrete and continuous latent variables but they are independent and $z$ follows an non-informative prior. These designs make it less powerful than ours in both generation quality and interpretability as evidenced in our experiments. 

\textbf{Energy-based model.\quad} Recent works \cite{xie2016theory, nijkamp2019learning, Han2020CVPR} demonstrate the effectiveness of EBMs in modeling complex dependency. \citet{pang2020learning} proposes to learn an EBM in the latent space as a prior model for the continuous latent vector, which greatly improves the model expressivity and demonstrates strong performance on text, image, molecule generation, and trajectory generation \cite{pang2020molecule, pang2021trajectory}. We also recruit an EBM as the prior model but this EBM couples a continuous vector and a discrete one, allowing for learning a more structured latent space, rendering generation interpretable, and admitting classification. In addition, the prior work uses MCMC for posterior inference but we recruits an inference network, $q_\phi(z|x)$, so that we can efficiently optimize over it, which is necessary for learning with the information bottleneck principle. Thus, this design admits a natural extension based on information bottleneck. 

\citet{grathwohl2019your} proposes the joint energy-based model (JEM) which is a classifier based EBM. Our model moves JEM to latent space. This brings two benefits. (1) Learning EBM in the data space usually involves expensive MCMC sampling. Our EBM is built in the latent space which has a much lower dimension and thus the sampling is much faster and has better mixing. (2) It is not straightforward to apply JEM to text data since it uses gradient-based sampling while the data space of text is non-differentiable.

\textbf{Information bottleneck.\quad} Information bottleneck proposed by \citet{tishby2000information} is an appealing principle to find good representations that trade-offs between the minimality of the representation and its sufficiency for predicting labels. Computing mutual information involved in applying this principle is however often computationally challenging. \citet{alemi2016deep} proposes a variational approach to reduce the computation complexity and uses it train supervised classifiers. In contrast, the information bottleneck in our model is embedded in a generative model and learned in an unsupervised manner. 

\section{Conclusion}
In this work, we formulate a latent space EBM which couples a dense vector for generation and a symbolic vector for interpretability and classification. The symbol or category can be inferred from the observed example based on the dense vector. The latent space EBM is used as the prior model for text generation model. The symbol-vector coupling, the generator network, and the inference network are learned jointly by maximizing the variational lower bound of the log-likelihood. Our model can be learned in unsupervised setting and the learning can be naturally extended to semi-supervised setting. The coupling formulation and the variational learning together naturally admit an incorporation of information bottleneck which encourages the continuous latent vector to extract information from the observed example that is informative of the underlying symbol. Our experiments demonstrate that the proposed model learns a well-structured and meaningful latent space, which (1) guides the top-down generator to generate text with high quality and interpretability, and (2) can be leveraged to effectively and accurately classify text.

\section*{Acknowledgments}
The work is partially supported by NSF DMS 2015577 and DARPA XAI project N66001-17-2-4029. We thank Erik Nijkamp and Tian Han for earlier collaborations.

\bibliography{example_paper}

\begin{thebibliography}{48}
\providecommand{\natexlab}[1]{#1}
\providecommand{\url}[1]{\texttt{#1}}
\expandafter\ifx\csname urlstyle\endcsname\relax
  \providecommand{\doi}[1]{doi: #1}\else
  \providecommand{\doi}{doi: \begingroup \urlstyle{rm}\Url}\fi

\bibitem[Alemi et~al.(2016)Alemi, Fischer, Dillon, and Murphy]{alemi2016deep}
Alemi, A.~A., Fischer, I., Dillon, J.~V., and Murphy, K.
\newblock Deep variational information bottleneck.
\newblock \emph{arXiv preprint arXiv:1612.00410}, 2016.

\bibitem[Aneja et~al.(2020)Aneja, Schwing, Kautz, and Vahdat]{aneja2020ncpvae}
Aneja, J., Schwing, A., Kautz, J., and Vahdat, A.
\newblock Ncp-vae: Variational autoencoders with noise contrastive priors,
  2020.

\bibitem[Bowman et~al.(2016)Bowman, Vilnis, Vinyals, Dai, Jozefowicz, and
  Bengio]{bowman-etal-2016-generating}
Bowman, S.~R., Vilnis, L., Vinyals, O., Dai, A., Jozefowicz, R., and Bengio, S.
\newblock Generating sentences from a continuous space.
\newblock In \emph{Proceedings of The 20th {SIGNLL} Conference on Computational
  Natural Language Learning}, pp.\  10--21, Berlin, Germany, August 2016.
  Association for Computational Linguistics.
\newblock \doi{10.18653/v1/K16-1002}.
\newblock URL \url{https://www.aclweb.org/anthology/K16-1002}.

\bibitem[Brown et~al.(1993)Brown, Della~Pietra, Della~Pietra, and
  Mercer]{brown1993mathematics}
Brown, P.~F., Della~Pietra, S.~A., Della~Pietra, V.~J., and Mercer, R.~L.
\newblock The mathematics of statistical machine translation: Parameter
  estimation.
\newblock \emph{Computational linguistics}, 19\penalty0 (2):\penalty0 263--311,
  1993.

\bibitem[Burda et~al.(2015)Burda, Grosse, and
  Salakhutdinov]{burda2015importance}
Burda, Y., Grosse, R., and Salakhutdinov, R.
\newblock Importance weighted autoencoders.
\newblock \emph{arXiv preprint arXiv:1509.00519}, 2015.

\bibitem[Card et~al.(2018)Card, Tan, and Smith]{card2018neural}
Card, D., Tan, C., and Smith, N.~A.
\newblock Neural models for documents with metadata.
\newblock In \emph{Proceedings of the 56th Annual Meeting of the Association
  for Computational Linguistics (Volume 1: Long Papers)}, pp.\  2031--2040,
  2018.

\bibitem[Cho et~al.(2014)Cho, van Merri{\"e}nboer, Gulcehre, Bahdanau,
  Bougares, Schwenk, and Bengio]{cho2014learning}
Cho, K., van Merri{\"e}nboer, B., Gulcehre, C., Bahdanau, D., Bougares, F.,
  Schwenk, H., and Bengio, Y.
\newblock Learning phrase representations using rnn encoder--decoder for
  statistical machine translation.
\newblock In \emph{Proceedings of the 2014 Conference on Empirical Methods in
  Natural Language Processing (EMNLP)}, pp.\  1724--1734, 2014.

\bibitem[Eric et~al.(2017)Eric, Krishnan, Charette, and
  Manning]{eric2017key-value}
Eric, M., Krishnan, L., Charette, F., and Manning, C.~D.
\newblock Key-value retrieval networks for task-oriented dialogue.
\newblock \emph{Annual Meeting of the Special Interest Group on Discourse and
  Dialogue}, pp.\  37--49, 2017.

\bibitem[Forgues et~al.(2014)Forgues, Pineau, Larchev{\^e}que, and
  Tremblay]{forgues2014bootstrapping}
Forgues, G., Pineau, J., Larchev{\^e}que, J.-M., and Tremblay, R.
\newblock Bootstrapping dialog systems with word embeddings.
\newblock In \emph{NIPS, Modern Machine Learning and Natural Language
  Processing Workshop}, volume~2, 2014.

\bibitem[Fu et~al.(2019)Fu, Li, Liu, Gao, Celikyilmaz, and
  Carin]{fu-etal-2019-cyclical}
Fu, H., Li, C., Liu, X., Gao, J., Celikyilmaz, A., and Carin, L.
\newblock Cyclical annealing schedule: A simple approach to mitigating {KL}
  vanishing.
\newblock In \emph{Proceedings of the 2019 Conference of the North {A}merican
  Chapter of the Association for Computational Linguistics: Human Language
  Technologies, Volume 1 (Long and Short Papers)}, pp.\  240--250, Minneapolis,
  Minnesota, June 2019. Association for Computational Linguistics.
\newblock \doi{10.18653/v1/N19-1021}.
\newblock URL \url{https://www.aclweb.org/anthology/N19-1021}.

\bibitem[Grathwohl et~al.(2019)Grathwohl, Wang, Jacobsen, Duvenaud, Norouzi,
  and Swersky]{grathwohl2019your}
Grathwohl, W., Wang, K.-C., Jacobsen, J.-H., Duvenaud, D., Norouzi, M., and
  Swersky, K.
\newblock Your classifier is secretly an energy based model and you should
  treat it like one.
\newblock In \emph{International Conference on Learning Representations}, 2019.

\bibitem[Gupta et~al.(2018)Gupta, Agarwal, Singh, and Rai]{gupta2018deep}
Gupta, A., Agarwal, A., Singh, P., and Rai, P.
\newblock A deep generative framework for paraphrase generation.
\newblock In \emph{Proceedings of the AAAI Conference on Artificial
  Intelligence}, volume~32, 2018.

\bibitem[Gururangan et~al.(2019)Gururangan, Dang, Card, and
  Smith]{gururangan2019variational}
Gururangan, S., Dang, T., Card, D., and Smith, N.~A.
\newblock Variational pretraining for semi-supervised text classification.
\newblock In \emph{Proceedings of the 57th Annual Meeting of the Association
  for Computational Linguistics}, pp.\  5880--5894, 2019.

\bibitem[Han et~al.(2020)Han, Nijkamp, Zhou, Pang, Zhu, and Wu]{Han2020CVPR}
Han, T., Nijkamp, E., Zhou, L., Pang, B., Zhu, S.-C., and Wu, Y.~N.
\newblock Joint training of variational auto-encoder and latent energy-based
  model.
\newblock In \emph{The IEEE/CVF Conference on Computer Vision and Pattern
  Recognition (CVPR)}, June 2020.

\bibitem[He et~al.(2019)He, Spokoyny, Neubig, and
  Berg-Kirkpatrick]{he2018lagging}
He, J., Spokoyny, D., Neubig, G., and Berg-Kirkpatrick, T.
\newblock Lagging inference networks and posterior collapse in variational
  autoencoders.
\newblock In \emph{Proceedings of ICLR}, 2019.

\bibitem[Jin et~al.(2020)Jin, Wiseman, Stratos, and
  Livescu]{jin-etal-2020-discrete}
Jin, S., Wiseman, S., Stratos, K., and Livescu, K.
\newblock Discrete latent variable representations for low-resource text
  classification.
\newblock In \emph{Proceedings of the 58th Annual Meeting of the Association
  for Computational Linguistics}, pp.\  4831--4842, Online, July 2020.
  Association for Computational Linguistics.
\newblock \doi{10.18653/v1/2020.acl-main.437}.
\newblock URL \url{https://www.aclweb.org/anthology/2020.acl-main.437}.

\bibitem[Kingma \& Welling(2014)Kingma and Welling]{kingma2013auto}
Kingma, D.~P. and Welling, M.
\newblock Auto-encoding variational bayes.
\newblock In \emph{2nd International Conference on Learning Representations,
  {ICLR} 2014, Banff, AB, Canada, April 14-16, 2014, Conference Track
  Proceedings}, 2014.
\newblock URL \url{http://arxiv.org/abs/1312.6114}.

\bibitem[Kingma et~al.(2014)Kingma, Mohamed, Rezende, and
  Welling]{kingma2014semi}
Kingma, D.~P., Mohamed, S., Rezende, D.~J., and Welling, M.
\newblock Semi-supervised learning with deep generative models.
\newblock In \emph{Advances in neural information processing systems}, pp.\
  3581--3589, 2014.

\bibitem[Li et~al.(2019)Li, He, Neubig, Berg-Kirkpatrick, and
  Yang]{li-etal-2019-surprisingly}
Li, B., He, J., Neubig, G., Berg-Kirkpatrick, T., and Yang, Y.
\newblock A surprisingly effective fix for deep latent variable modeling of
  text.
\newblock In \emph{Proceedings of the 2019 Conference on Empirical Methods in
  Natural Language Processing and the 9th International Joint Conference on
  Natural Language Processing (EMNLP-IJCNLP)}, pp.\  3603--3614, Hong Kong,
  China, November 2019. Association for Computational Linguistics.
\newblock \doi{10.18653/v1/D19-1370}.
\newblock URL \url{https://www.aclweb.org/anthology/D19-1370}.

\bibitem[Li et~al.(2018)Li, Jia, He, and Liang]{li-etal-2018-delete}
Li, J., Jia, R., He, H., and Liang, P.
\newblock Delete, retrieve, generate: a simple approach to sentiment and style
  transfer.
\newblock In \emph{Proceedings of the 2018 Conference of the North {A}merican
  Chapter of the Association for Computational Linguistics: Human Language
  Technologies, Volume 1 (Long Papers)}, pp.\  1865--1874, New Orleans,
  Louisiana, June 2018. Association for Computational Linguistics.
\newblock \doi{10.18653/v1/N18-1169}.
\newblock URL \url{https://www.aclweb.org/anthology/N18-1169}.

\bibitem[Li et~al.(2017{\natexlab{a}})Li, Lam, Bing, and Wang]{li2017deep}
Li, P., Lam, W., Bing, L., and Wang, Z.
\newblock Deep recurrent generative decoder for abstractive text summarization.
\newblock In \emph{Proceedings of the 2017 Conference on Empirical Methods in
  Natural Language Processing}, pp.\  2091--2100, 2017{\natexlab{a}}.

\bibitem[Li et~al.(2017{\natexlab{b}})Li, Su, Shen, Li, Cao, and
  Niu]{li2017dailydialog}
Li, Y., Su, H., Shen, X., Li, W., Cao, Z., and Niu, S.
\newblock Dailydialog: A manually labelled multi-turn dialogue dataset.
\newblock \emph{International Joint Conference on Natural Language Processing},
  1:\penalty0 986--995, 2017{\natexlab{b}}.

\bibitem[Marcus et~al.(1993)Marcus, Marcinkiewicz, and
  Santorini]{marcus1993building}
Marcus, M.~P., Marcinkiewicz, M.~A., and Santorini, B.
\newblock Building a large annotated corpus of english: The penn treebank.
\newblock \emph{Comput. Linguist.}, 19\penalty0 (2):\penalty0 313--330, June
  1993.
\newblock ISSN 0891-2017.
\newblock URL \url{http://dl.acm.org/citation.cfm?id=972470.972475}.

\bibitem[Merity et~al.(2018)Merity, Keskar, and Socher]{merity2018regularizing}
Merity, S., Keskar, N.~S., and Socher, R.
\newblock Regularizing and optimizing lstm language models.
\newblock In \emph{International Conference on Learning Representations}, 2018.

\bibitem[Miao et~al.(2016)Miao, Yu, and Blunsom]{miao2016neural}
Miao, Y., Yu, L., and Blunsom, P.
\newblock Neural variational inference for text processing.
\newblock In \emph{International conference on machine learning}, pp.\
  1727--1736. PMLR, 2016.

\bibitem[Mikolov et~al.(2010)Mikolov, Karafi{\'a}t, Burget, {\v{C}}ernock{\`y},
  and Khudanpur]{mikolov2010recurrent}
Mikolov, T., Karafi{\'a}t, M., Burget, L., {\v{C}}ernock{\`y}, J., and
  Khudanpur, S.
\newblock Recurrent neural network based language model.
\newblock In \emph{Eleventh annual conference of the international speech
  communication association}, 2010.

\bibitem[Mitchell \& Lapata(2008)Mitchell and Lapata]{mitchell2008vector}
Mitchell, J. and Lapata, M.
\newblock Vector-based models of semantic composition.
\newblock \emph{proceedings of ACL-08: HLT}, pp.\  236--244, 2008.

\bibitem[Nijkamp et~al.(2019)Nijkamp, Hill, Zhu, and Wu]{nijkamp2019learning}
Nijkamp, E., Hill, M., Zhu, S.-C., and Wu, Y.~N.
\newblock Learning non-convergent non-persistent short-run {MCMC} toward
  energy-based model.
\newblock \emph{Advances in Neural Information Processing Systems 33: Annual
  Conference on Neural Information Processing Systems 2019, NeurIPS 2019, 8-14
  December 2019, Vancouver, Canada}, 2019.

\bibitem[Pang et~al.(2020{\natexlab{a}})Pang, Han, Nijkamp, Zhu, and
  Wu]{pang2020learning}
Pang, B., Han, T., Nijkamp, E., Zhu, S.-C., and Wu, Y.~N.
\newblock Learning latent space energy-based prior model.
\newblock \emph{Advances in Neural Information Processing Systems}, 33,
  2020{\natexlab{a}}.

\bibitem[Pang et~al.(2020{\natexlab{b}})Pang, Han, and Wu]{pang2020molecule}
Pang, B., Han, T., and Wu, Y.~N.
\newblock Learning latent space energy-based prior model for molecule
  generation.
\newblock \emph{arXiv preprint arXiv:2010.09351}, 2020{\natexlab{b}}.

\bibitem[Pang et~al.(2021)Pang, Zhao, Xie, and Wu]{pang2021trajectory}
Pang, B., Zhao, T., Xie, X., and Wu, Y.~N.
\newblock Trajectory prediction with latent belief energy-based model.
\newblock \emph{arXiv preprint arXiv:2104.03086}, 2021.

\bibitem[Papineni et~al.(2002)Papineni, Roukos, Ward, and
  Zhu]{papineni2002bleu}
Papineni, K., Roukos, S., Ward, T., and Zhu, W.-J.
\newblock Bleu: a method for automatic evaluation of machine translation.
\newblock In \emph{Proceedings of the 40th annual meeting of the Association
  for Computational Linguistics}, pp.\  311--318, 2002.

\bibitem[Rezende et~al.(2014)Rezende, Mohamed, and
  Wierstra]{rezende2014stochastic}
Rezende, D.~J., Mohamed, S., and Wierstra, D.
\newblock Stochastic backpropagation and approximate inference in deep
  generative models.
\newblock In \emph{Proceedings of the 31th International Conference on Machine
  Learning, {ICML} 2014, Beijing, China, 21-26 June 2014}, pp.\  1278--1286,
  2014.
\newblock URL \url{http://proceedings.mlr.press/v32/rezende14.html}.

\bibitem[Rus \& Lintean(2012)Rus and Lintean]{rus2012comparison}
Rus, V. and Lintean, M.
\newblock A comparison of greedy and optimal assessment of natural language
  student input using word-to-word similarity metrics.
\newblock In \emph{Proceedings of the Seventh Workshop on Building Educational
  Applications Using NLP}, pp.\  157--162. Association for Computational
  Linguistics, 2012.

\bibitem[Serban et~al.(2016)Serban, Sordoni, Bengio, Courville, and
  Pineau]{serban2016building}
Serban, I., Sordoni, A., Bengio, Y., Courville, A., and Pineau, J.
\newblock Building end-to-end dialogue systems using generative hierarchical
  neural network models.
\newblock In \emph{Proceedings of the AAAI Conference on Artificial
  Intelligence}, volume~30, 2016.

\bibitem[Serban et~al.(2017)Serban, Sordoni, Lowe, Charlin, Pineau, Courville,
  and Bengio]{serban2017hierarchical}
Serban, I., Sordoni, A., Lowe, R., Charlin, L., Pineau, J., Courville, A., and
  Bengio, Y.
\newblock A hierarchical latent variable encoder-decoder model for generating
  dialogues.
\newblock In \emph{Proceedings of the AAAI Conference on Artificial
  Intelligence}, volume~31, 2017.

\bibitem[Shi et~al.(2020)Shi, Zhou, Miao, and Li]{shi2020dispersed}
Shi, W., Zhou, H., Miao, N., and Li, L.
\newblock Dispersed exponential family mixture vaes for interpretable text
  generation.
\newblock In \emph{International Conference on Machine Learning}, pp.\
  8840--8851. PMLR, 2020.

\bibitem[Subramanian et~al.(2018)Subramanian, Rajeswar, Sordoni, Trischler,
  Courville, and Pal]{subramanian2018towards}
Subramanian, S., Rajeswar, S., Sordoni, A., Trischler, A., Courville, A., and
  Pal, C.
\newblock Towards text generation with adversarially learned neural outlines.
\newblock In \emph{Proceedings of the 32nd International Conference on Neural
  Information Processing Systems}, pp.\  7562--7574, 2018.

\bibitem[Tishby et~al.(2000)Tishby, Pereira, and Bialek]{tishby2000information}
Tishby, N., Pereira, F.~C., and Bialek, W.
\newblock The information bottleneck method.
\newblock \emph{arXiv preprint physics/0004057}, 2000.

\bibitem[Vincent et~al.(2010)Vincent, Larochelle, Lajoie, Bengio, and
  Manzagol]{vincent2010stacked}
Vincent, P., Larochelle, H., Lajoie, I., Bengio, Y., and Manzagol, P.~A.
\newblock Stacked denoising autoencoders: Learning useful representations in a
  deep network with a local denoising criterion.
\newblock \emph{Journal of Machine Learning Research}, 11\penalty0
  (12):\penalty0 3371--3408, 2010.

\bibitem[Wang et~al.(2019)Wang, Gan, Xu, Zhang, Wang, Shen, Chen, and
  Carin]{wang2019topic}
Wang, W., Gan, Z., Xu, H., Zhang, R., Wang, G., Shen, D., Chen, C., and Carin,
  L.
\newblock Topic-guided variational auto-encoder for text generation.
\newblock In \emph{Proceedings of the 2019 Conference of the North American
  Chapter of the Association for Computational Linguistics: Human Language
  Technologies, Volume 1 (Long and Short Papers)}, pp.\  166--177, 2019.

\bibitem[Xie et~al.(2016)Xie, Lu, Zhu, and Wu]{xie2016theory}
Xie, J., Lu, Y., Zhu, S., and Wu, Y.~N.
\newblock A theory of generative convnet.
\newblock In \emph{Proceedings of the 33nd International Conference on Machine
  Learning, {ICML} 2016, New York City, NY, USA, June 19-24, 2016}, pp.\
  2635--2644, 2016.
\newblock URL \url{http://proceedings.mlr.press/v48/xiec16.html}.

\bibitem[Young et~al.(2013)Young, Ga{\v{s}}i{\'c}, Thomson, and
  Williams]{young2013pomdp}
Young, S., Ga{\v{s}}i{\'c}, M., Thomson, B., and Williams, J.~D.
\newblock Pomdp-based statistical spoken dialog systems: A review.
\newblock \emph{Proceedings of the IEEE}, 101\penalty0 (5):\penalty0
  1160--1179, 2013.

\bibitem[Zhang et~al.(2016)Zhang, Xiong, Su, Duan, and
  Zhang]{zhang-etal-2016-variational-neural}
Zhang, B., Xiong, D., Su, J., Duan, H., and Zhang, M.
\newblock Variational neural machine translation.
\newblock In \emph{Proceedings of the 2016 Conference on Empirical Methods in
  Natural Language Processing}, pp.\  521--530, Austin, Texas, November 2016.
  Association for Computational Linguistics.
\newblock \doi{10.18653/v1/D16-1050}.
\newblock URL \url{https://www.aclweb.org/anthology/D16-1050}.

\bibitem[Zhang et~al.(2015)Zhang, Zhao, and LeCun]{zhang2015character}
Zhang, X., Zhao, J., and LeCun, Y.
\newblock Character-level convolutional networks for text classification.
\newblock In \emph{Advances in neural information processing systems}, pp.\
  649--657, 2015.

\bibitem[Zhao et~al.(2018{\natexlab{a}})Zhao, Kim, Zhang, Rush, and
  LeCun]{zhao2018adversarially}
Zhao, J., Kim, Y., Zhang, K., Rush, A., and LeCun, Y.
\newblock Adversarially regularized autoencoders.
\newblock In \emph{International Conference on Machine Learning}, pp.\
  5902--5911, 2018{\natexlab{a}}.

\bibitem[Zhao et~al.(2017)Zhao, Zhao, and Eskenazi]{zhao2017learning}
Zhao, T., Zhao, R., and Eskenazi, M.
\newblock Learning discourse-level diversity for neural dialog models using
  conditional variational autoencoders.
\newblock In \emph{Proceedings of the 55th Annual Meeting of the Association
  for Computational Linguistics (Volume 1: Long Papers)}, pp.\  654--664, 2017.

\bibitem[Zhao et~al.(2018{\natexlab{b}})Zhao, Lee, and
  Eskenazi]{zhao2018unsupervised}
Zhao, T., Lee, K., and Eskenazi, M.
\newblock Unsupervised discrete sentence representation learning for
  interpretable neural dialog generation.
\newblock In \emph{Proceedings of the 56th Annual Meeting of the Association
  for Computational Linguistics (Volume 1: Long Papers)}, pp.\  1098--1107,
  2018{\natexlab{b}}.

\end{thebibliography}
\bibliographystyle{icml2021}





\end{document}